\pdfoutput=1
\documentclass[journal]{IEEEtran}
\usepackage[T1]{fontenc}
\usepackage{cite}
\usepackage{amsmath,amssymb,amsfonts,bm}
\usepackage{graphicx}
\usepackage{textcomp}
\usepackage{booktabs}
\usepackage{siunitx}
\usepackage{tikz}
\usetikzlibrary{arrows.meta,positioning,shapes.geometric,fit,calc,backgrounds}
\usepackage{microtype}
\usepackage{multirow}
\usepackage{placeins}
\usepackage[hidelinks]{hyperref}
\hypersetup{pdftitle={RTISSLAM-3D: Full 6-DoF SLAM With an In-Air 3D Imaging Sonar},
  pdfauthor={Jan Steckel, Vincent Katter, Wouter Jansen}}

\begin{document}

\title{RTISSLAM-3D: Full 6-DoF SLAM With an\\ In-Air 3D Imaging Sonar}

\author{Jan~Steckel\textsuperscript{1,2}, Vincent~Katter\textsuperscript{1,2}, and Wouter~Jansen\textsuperscript{1,2}\\[4pt]
{\normalsize\textsuperscript{1}Cosys-Lab, Faculty of Applied Engineering, University of Antwerp, 2020 Antwerp, Belgium\\
\textsuperscript{2}Flanders Make Strategic Research Centre, 3920 Lommel, Belgium\\
e-mail: jan.steckel@uantwerpen.be}%
\thanks{This work has been submitted to the IEEE for possible publication. Copyright may be transferred without notice, after which this version may no longer be accessible.}%
\thanks{A supplementary video with the raw sonar data, the SLAM being built, and a drive through the sonar and lidar 3D maps is available at \url{https://youtu.be/GHdZd9Brqog} and as an ancillary file of this arXiv submission.}}

\markboth{Preprint}%
{Steckel \MakeLowercase{\textit{et al.}}: RTISSLAM-3D: Full 6-DoF SLAM With an In-Air 3D Imaging Sonar}

\maketitle

\begin{abstract}
Echolocating bats navigate dense forests using sonar alone, and in-air 3D imaging sonars now capture the entire frontal hemisphere with every pulse. Sonar SLAM, however, has so far been planar: current systems estimate a 2D trajectory and discard the elevation information in the sonar images, an assumption that does not hold on rough and sloped terrain such as a forest floor. In this paper, we introduce RTISSLAM-3D, a novel 6-DoF SLAM system for a wheeled robot that carries an in-air 3D imaging sonar, an IMU and wheel encoders. A CFAR detector on the sphere reduces every sonar image to a sparse set of 3D detections, which we use for a 3D acoustic odometry that models the shape of the sonar beam, for place recognition with polar submap descriptors, and for the verification of loop closures by 3D registration, all combined in a 6-DoF pose graph with attitude factors. We evaluated RTISSLAM-3D on a \SI{545}{\meter} drive through a forest, with a 3D lidar SLAM as reference. The loop closures reduced the horizontal error of the acoustic odometry from \SI{1.74}{\meter} to \SI{0.28}{\meter}, with a height error of \SI{0.12}{\meter}, and \SI{94.6}{\percent} of the resulting 3D sonar map lies within \SI{0.3}{\meter} of the lidar map.

\end{abstract}


\section{Introduction}
\label{sec:intro}
Sonar was one of the first sensors with which mobile robots built maps \cite{moravecHighResolutionMaps1985, elfesUsingOccupancyGrids1989, leonardMobileRobotLocalization1991}, and in-air sonar keeps working in exactly those conditions in which cameras and lidar struggle: in darkness, smoke, dust and fog, and next to glass \cite{steckelBroadband3DSonar2013, kerstensERTISFullyEmbedded2019, Laurijssen2026eRTIS}. Indeed, ultrasonic sensors are cheap, small and robust, and the embedded 3D imaging sonars of the last decade, such as the eRTIS \cite{kerstensERTISFullyEmbedded2019, Laurijssen2026eRTIS}, turn the echoes of a single broadband pulse into an image of the reflected energy as a function of range and direction over the complete frontal hemisphere, the so-called energyscape \cite{steckelBroadband3DSonar2013}. With nonlinear beamformers such as Delay-Multiply-And-Sum (DMAS) with a coherence factor \cite{Jansen2026DMAS}, these images have become sharp enough to be used for more than obstacle avoidance and reactive control \cite{jansenRealTimeSonarFusion2022, steckelAcousticFlowBasedControl2017}.

In two recent papers, we have shown how such sonar data can drive a modern, robust SLAM back-end. BatSLAM 2.0 \cite{steckelBatSLAM20Sequenceverified2026} combined sequence-verified place recognition on the echoes of a binaural bat-like sonar with a factor-graph back-end \cite{kaessISAM2IncrementalSmoothing2012, dellaertFactorGraphsGTSAM2012}, and produced reliable topological maps in strongly aliased environments. RTISSLAM \cite{steckelRTISSLAMInAirImaging2026} replaced the binaural sensor by an eRTIS imaging sonar and used its energyscapes as place descriptors. Furthermore, the system extracts acoustic odometry from the sequence of the data, and anchored an occupancy grid to the nodes of the pose graph, so that a metric map is constructed, updated by every loop closure. Both systems, however, are operating in a strict 2D assumption: they estimate the position and heading of the robot on a plane, and they use the 3D imaging capability of the sensor only to build a 2D map, by suppressing the elevation dimension of the energyscape (looking at a horizontal slice). This is a reasonable choice in an office or a warehouse with flat floors, but it throws away a significant portion of what the sensor measures. Furthermore, this approach breaks down as soon as the robot leaves the flat floor, as is common in offroad scenarios, and for UAV navigation.

Outdoors, and in particular in a forest, the ground is neither flat nor level, and the robot rolls and pitches as it drives over roots and stones. The relevant structure (tree trunks, shrubs, fallen branches) is best represented in 3D, and a planar slice through it at the height of the sensor depends on the attitude of the robot. Furthermore, a forest is a notoriously difficult environment for place recognition as it can be strongly self-similar. In our own first experiments with the 2D RTISSLAM on real forest data, the recall of single-view place recognition at a useful precision was close to zero, and loop closures could only be found by restricting the search to the vicinity of the current pose estimate. 

In lidar SLAM, the step from 2D to 3D has long been made: 3D point clouds are registered with point-to-point or point-to-distribution ICP variants \cite{segalGeneralizedICP2009, biberNormalDistributionsTransform2003, vizzoKISSICPDefensePointtoPoint2023}, places are recognized with egocentric polar descriptors of the 3D structure \cite{kimScanContextEgocentric2018}, and maps are stored as 3D occupancy grids \cite{hornungOctoMapEfficientProbabilistic2013}. Underwater, multibeam and imaging sonars have been used to build volumetric submaps \cite{teixeiraUnderwaterInspectionUsing2016}. For in-air sonar, however, a full 3D SLAM system, estimating the 6-DoF trajectory of the robot and a 3D map from the sonar, has, to the best of our knowledge, not been demonstrated on a real robot. The data of an in-air imaging sonar is quite different from a lidar scan: a capture yields at most a few dozen detections within a range of a few meters, the angular resolution is a few degrees, often with anisotropy in azimuth and elevation, specular surfaces are only seen near normal incidence, and the ground returns both a direct echo and a mirror image through ground multipath. A sonar SLAM system needs to take all of these peculiarities into account when trying to succesfully build a spatial map.

In this paper, we propose RTISSLAM-3D, a novel full 6-DoF SLAM system for an in-air 3D imaging sonar on a wheeled robot, which uses the sonar's 3D images for every part of the localization: for odometry, for place recognition and for the metric verification of loop closures. Only an IMU (Inertial Measurement Unit) and the wheel encoders complement the sonar, and no learned components are used in the whole pipeline. In our approach, every capture is imaged over the full frontal hemisphere and turned into a sparse set of 3D detections by a CFAR detector that operates on the sphere, efficiently implemented through the heat kernel of a graph Laplacian, which is in effect a Gaussian blur over a set of irregular directions on a sphere. The ground echoes give the height and tilt of the sensor above the local ground and reveal the multipath mirror images, which are removed. A 3D acoustic odometry registers every detection set to a local 3D submap, with the beam-shaped uncertainty of each detection, and takes the rotation from the gyroscope. Keyframes enter a 6-DoF pose graph with IMU attitude factors, loop closures are proposed by a polar descriptor of the 3D submap and verified by a 4-DoF registration of the two submaps, and a 3D log-odds occupancy map is built on the optimized poses.

We evaluate RTISSLAM-3D on a \SI{545}{\meter} drive through a forest in Trondheim, Norway, with a gravity-aligned 3D lidar SLAM as reference, and we show that it reduces the horizontal trajectory error of the wheel and gyroscope odometry from \SI{2.40}{\meter} to \SI{0.28}{\meter}, with a height error of \SI{0.12}{\meter}, closes 233 loops of which none is off by more than \SI{0.5}{\meter}, and builds a 3D map in which \SI{94.6}{\percent} of the occupied voxels lie within \SI{0.3}{\meter} of the lidar ground-truth map.

The contributions of this paper are as follows: (1) a 3D imaging and detection chain for in-air sonar over the full frontal hemisphere, with a CFAR detector on the sphere and the removal of ground multipath through a fitted ground plane, (2) a 3D acoustic odometry that registers sparse sonar detections with their beam-shaped uncertainty, (3) 3D place recognition for sonar, with candidates proposed by a polar submap descriptor and verified by submap registration, (4) the complete 6-DoF SLAM system and a 3D occupancy map with beam-shaped evidence and free-space carving, and (5) an evaluation on real data against a gravity-aligned 3D lidar reference, with ablations of every component. The complete pipeline, from the raw recording to the map and the videos, runs with a single command and will be made available as open source.

The rest of the paper is structured as follows. First, we give an overview of RTISSLAM-3D in section \ref{sec:overview}. Next, we detail the imaging and detection chain, the acoustic odometry, the place recognition, the pose graph and the mapping in section \ref{sec:method}. Section \ref{sec:setup} describes the robot, the recording and the reference, after which section \ref{sec:results} presents the results. We discuss the limitations of our work in section \ref{sec:discussion} and conclude the paper in section \ref{sec:conclusion}.

\section{Overview of RTISSLAM-3D}
\label{sec:overview}
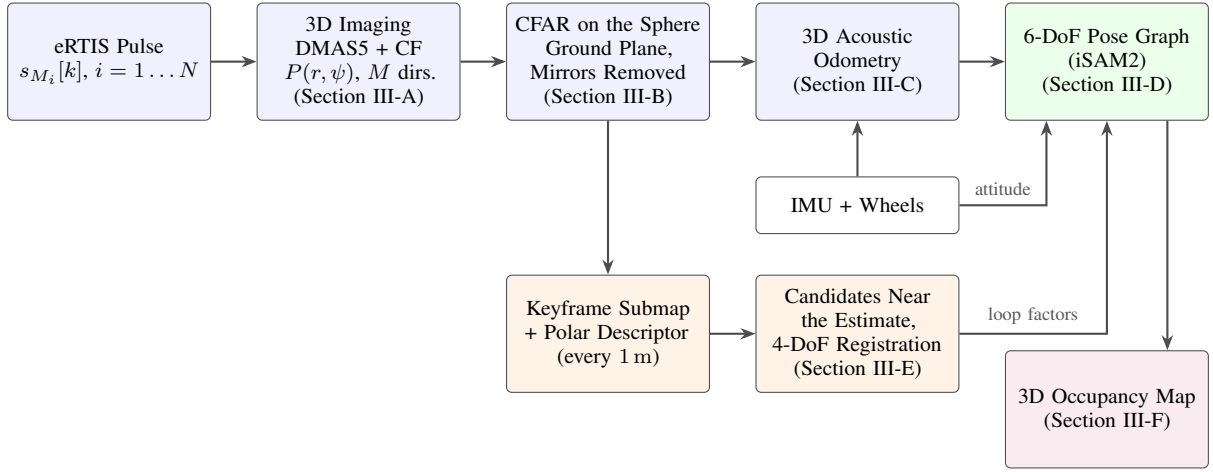
\begin{figure*}[t]
\centering
\begin{tikzpicture}[
  box/.style={draw=black!70, rounded corners=2pt, align=center, font=\footnotesize, text width=2.45cm, minimum height=#1},
  box/.default=1.55cm,
  arr/.style={-{Stealth[length=2mm]}, thick, black!70},
  lab/.style={font=\scriptsize, black!65, align=center}]
\node[box, fill=blue!6] (echo) at (0,0) {eRTIS Pulse\\ $s_{M_i}[k]$, $i = 1 \ldots N$};
\node[box, fill=blue!6] (img) at (3.3,0) {3D Imaging\\ DMAS5 + CF\\ $P(r,\psi)$, $M$ dirs.\\ (Section \ref{sec:imaging})};
\node[box, fill=blue!6] (det) at (6.6,0) {CFAR on the Sphere\\ Ground Plane,\\ Mirrors Removed\\ (Section \ref{sec:detections})};
\node[box, fill=blue!6] (ao) at (9.9,0) {3D Acoustic\\ Odometry\\ (Section \ref{sec:ao})};
\node[box, fill=green!8] (pg) at (13.2,0) {6-DoF Pose Graph\\ (iSAM2)\\ (Section \ref{sec:posegraph})};
\node[box=0.75cm, fill=white] (imu) at (9.9,-1.9) {IMU + Wheels};
\node[box, fill=orange!10] (kf) at (6.6,-3.6) {Keyframe Submap\\ + Polar Descriptor\\ (every \SI{1}{\meter})};
\node[box, fill=orange!10] (pr) at (9.9,-3.6) {Candidates Near\\ the Estimate,\\ 4-DoF Registration\\ (Section \ref{sec:place})};
\node[box, fill=purple!8] (map) at (13.2,-4.6) {3D Occupancy Map\\ (Section \ref{sec:map})};
\draw[arr] (echo) -- (img);
\draw[arr] (img) -- (det);
\draw[arr] (det) -- (ao);
\draw[arr] (ao) -- (pg);
\draw[arr] (imu) -- (ao);
\draw[arr] (imu.east) -| node[lab, pos=0.25, above] {attitude} ([xshift=-8mm]pg.south);
\draw[arr] (det) -- (kf);
\draw[arr] (kf) -- (pr);
\draw[arr] (pr.east) -| node[lab, pos=0.25, above] {loop factors} (pg.south);
\draw[arr] ([xshift=8mm]pg.south) -- ([xshift=8mm]map.north);
\end{tikzpicture}
\caption{Overview of RTISSLAM-3D. Every eRTIS pulse is imaged over the frontal hemisphere (DMAS5 with coherence factor) and reduced to a sparse 3D detection set by a CFAR detector on the sphere; a fitted ground plane removes the ground multipath mirrors. The detection sets drive a 3D acoustic odometry (sonar translation, gyroscope rotation), keyframes every \SI{1}{\meter} enter a 6-DoF pose graph with IMU attitude factors, and loop closures are proposed by a polar descriptor of the keyframe submaps and verified by 4-DoF submap registration. The 3D occupancy map is built on the optimized poses.}
\label{fig:overview}
\end{figure*}

Figure \ref{fig:overview} shows the processing flow of RTISSLAM-3D. For every pulse, the eRTIS records the echoes with its $N = 32$ microphones, which are beamformed with fifth-order DMAS with coherence factor (CF) into an energy volume $P(r,\psi)$ over $M = 2293$ directions $\psi$ that sample the frontal hemisphere uniformly (section \ref{sec:imaging}). A CFAR detector on the sphere and a peak threshold turn every volume into a sparse set of 3D detections, after which a ground plane is fitted to the near-field ground echoes and the multipath mirror images below it are removed (section \ref{sec:detections}). Each detection set is registered to a local submap of the previous captures, with the wheel distance and the gyroscope rotation as prior, which yields the 3D acoustic odometry (section \ref{sec:ao}). After every \SI{1}{\meter} of travel, a keyframe is created: a 6-DoF node in the pose graph, connected to the previous keyframe by the acoustic odometry and to the direction of gravity by an IMU attitude factor (section \ref{sec:posegraph}). The keyframe also stores its submap, the detections of the last \SI{4}{\meter} of travel expressed in a level frame, together with a polar descriptor of that submap. Keyframes that were visited at least \SI{20}{\meter} of travel earlier and lie close to the current estimate are ranked by their descriptor, and the best candidates are verified by registering the two submaps; a verified revisit becomes a robust loop closure factor (section \ref{sec:place}). Finally, a 3D log-odds occupancy map is built from the detections on the optimized poses (section \ref{sec:map}). As in RTISSLAM \cite{steckelRTISSLAMInAirImaging2026}, the information flows in one direction: the map is built from the pose graph, but map information never feeds back into the pose-graph construction, which is a deliberate choice based on the peculiar reflection dynamics of ultrasound sensing \cite{steckelRTISSLAMInAirImaging2026}.

\section{Method}
\label{sec:method}
In this section, we detail the components of RTISSLAM-3D in the order of figure \ref{fig:overview}. We use the notation of RTISSLAM \cite{steckelRTISSLAMInAirImaging2026}, extended to 3D: the eRTIS has $N$ microphones at positions $p_y(i)$, $p_z(i)$ in its $YZ$-plane, and a direction $\psi$ has azimuth $\theta$ and elevation $\varphi$, with unit vector $\vec{u}(\psi)$ in the sensor frame ($x$ forward, $y$ left, $z$ up).

\subsection{3D Imaging Over the Frontal Hemisphere}
\label{sec:imaging}
The planar array of the eRTIS cannot distinguish echoes arriving front from back, so the frontal hemisphere holds all the directional information of a pulse. In contrast to RTISSLAM, which beamformed a single horizontal fan of directions, we image the complete hemisphere in this paper. To give every direction the same weight in the detector and in the maps that follow, the directions should sample the hemisphere uniformly in solid angle instead of uniformly in azimuth and elevation. We sample the frontal hemisphere with a Fibonacci (golden-angle) lattice \cite{gonzalezMeasurementAreasSphere2010}, keeping the points with a positive $x$-component: for a nominal spacing of \SI{3}{\degree}, this yields $M = 2293$ directions $\vec{u}_m$, each covering the same solid angle up to $O(1/M)$.

For every direction $\psi$, the matched-filtered microphone signals $s_{MF_i}[k]$ (the microphone spectra $S_{M_i}(j\omega)$ multiplied by the conjugate $S_b^*(j\omega)$ of the emitted call) are delayed by
\begin{equation}
\tau_i(\psi) = \frac{f_s}{c}\,\big(\psi_y\, p_y(i) + \psi_z\, p_z(i)\big),
\end{equation}
where $\psi_y$ and $\psi_z$ are the $y$- and $z$-components of the unit vector $\vec{u}_m$ of direction $\psi$, and $p_y(i)$, $p_z(i)$ the position of microphone $i$ in the array plane. The delayed signals are combined by fifth-order DMAS with coherence factor into the beamformed signal $s^{BF}_{\psi}[k]$ \cite{Jansen2026DMAS}. Its envelope $s^{EN}_{\psi}[k]$ (zero-phase low-pass filtered at \SI{5}{\kilo\hertz}) is sampled in range cells of \SI{2}{\centi\meter} up to \SI{9.7}{\meter}, which yields the energy volume $P(r,\psi)$ of dimensions $[475 \times M]$ (475 range cells). Before beamforming, we apply the same preprocessing as for the 2D RTISSLAM on real data: the direct path is aligned, a static background is subtracted, and the broadband ultrasonic interference of the robot's drive electronics is projected out: per frequency, the dominant eigenvectors of the cross-spectral matrix of the echo-poor part of the records (eigenvalues larger than four times the median, at most four) span an interference subspace, whose projector is removed from the microphone spectra.

\subsection{Detections: CFAR on the Sphere and Ground Multipath}
\label{sec:detections}
RTISSLAM used a range-wise CFAR (RW-CFAR) to detect targets and suppress sidelobes: for every range, the energy was compared with its mean over all directions. Over a full hemisphere, a mean over all directions is a poor model of the local clutter level: the ground fills the lower half of the hemisphere at short range, and a few trunks occupy small parts of it. We therefore estimate the background of every cell as a local weighted mean over neighboring directions, and we take inspiration from the heat equation on the sphere to define what "local" means on a non-regular set of directions. The heat kernel $e^{-t\Delta_{S^2}}$ of the Laplace-Beltrami operator is the natural local mean on the sphere. On our sampled hemisphere, we replace it by the heat kernel of a graph Laplacian \cite{belkinLaplacianEigenmaps2003, coifmanDiffusionMaps2006} for efficient implementation. We build a graph that connects every direction to its $k = 10$ nearest neighbors, with Gaussian weights $W_{mn}$ of a beam-matched angular distance in which elevation differences are divided by 1.7 (the main lobe of the eRTIS is about 1.7 times wider in elevation than in azimuth). With the degree matrix $D = \mathrm{diag}(\sum_n W_{mn})$, the random-walk Laplacian and its heat kernel are given by:
\begin{equation}
L = I - D^{-1}\cdot W, \qquad H_t = e^{-t\cdot L},
\end{equation}
where $H_t$ has dimensions $[M \times M]$ and is row-stochastic, so that it preserves constants and needs no special treatment at the edge of the hemisphere. We choose $t$ such that the RMS angular radius of the kernel at broadside equals \SI{12}{\degree}. The background $B(r,\psi)$ is the heat-kernel mean over directions, smoothed along range with a Gaussian of three range cells, and the CFAR output is:
\begin{equation}
P^{+}(r,\psi) = \max\big(P(r,\psi) - g\cdot B(r,\psi),\, 0\big),
\label{eq:cfar}
\end{equation}
with gain $g = 5$ in the experiments in this paper. 

For $t \to \infty$, $H_t$ converges to a (degree-weighted) mean over all directions, and equation \ref{eq:cfar} becomes the RW-CFAR of RTISSLAM. For small $t$, it collapses to a well-known cell-averaging CFAR on the sphere. The detections of a capture are the cells within \SI{25}{\decibel} of the strongest CFAR output, between \SI{0.5}{\meter} and \SI{8}{\meter}, each with its position $\vec{p}_j = r_j\cdot\vec{u}(\psi_j)$ in the sensor frame and its CFAR excess $w_j$ as weight.

The ground returns a direct echo at short range below the sensor, but also a mirror image of every object above it, through the path that bounces off the ground: in our point-target simulations, these mirrors are only 4 to \SI{9}{\decibel} weaker than the direct echoes. We remove them with a ground plane $\vec{n}^T\cdot\vec{x} = d$ that is fitted to the low near-field detections (closer than \SI{3}{\meter}, below the sensor) by iteratively reweighted least squares with a Huber weight, and accept it when its normal is within \SI{30}{\degree} of the vertical. Detections more than \SI{0.15}{\meter} below the plane are mirrors and are dropped. As a by-product, the plane gives the height of the sensor above the local ground and its tilt with respect to that ground.

\subsection{3D Acoustic Odometry}
\label{sec:ao}
Acoustic flow, the apparent motion of reflectors in the sonar images of a moving sensor, encodes the motion of the sensor itself and has been used for control and navigation with 3D sonar sensors \cite{steckelAcousticFlowBasedControl2017, jansenAdaptiveAcousticFlowBased2021}. RTISSLAM \cite{steckelRTISSLAMInAirImaging2026} used this principle as a source for planar odometry, by correlating consecutive 2D energyscapes and finding the optimal rotation and translation between the energyscapes. In 3D, we instead register the detections of every capture to a local submap, which holds the detections of the previous 20 captures, placed with their current pose estimates. Table~\ref{tab:ao} lists all parameters.

For in-air sonar, the range of a detection is precise, but its direction is only known to within the width of the beam, so that its uncertainty across the beam (in cartesian space) grows linearly with range. We therefore give every detection a covariance in the sensor frame that is elongated across the beam:
\begin{equation}
B_j = E_j
\begin{bmatrix}
\sigma_r^2 & 0 & 0 \\
0 & (r_j\,\sigma_\theta)^2 & 0 \\
0 & 0 & (r_j\,\sigma_\varphi)^2
\end{bmatrix}
E_j^T,
\end{equation}
where $r_j$ is the range of the detection, and the columns of the rotation $E_j$ are the direction of the detection and the unit vectors in azimuth and elevation perpendicular to it. The elevation term is the largest, as the beam of the eRTIS is wider in elevation than in azimuth. Next, the pose of the current capture is the predicted pose, corrected by a small motion $\vec{\xi}$ in the body frame:
\begin{equation}
T(\vec{\xi}) = T_{\text{pred}}\cdot\mathrm{Exp}(\vec{\xi}).
\end{equation}
Every detection is transformed into the submap and associated with the mean of its nearest submap points:
\begin{equation}
\vec{q}_j = T(\vec{\xi})\cdot\vec{p}_j, \qquad \vec{\mu}_j = \frac{1}{k}\sum_{n \in \mathcal{N}_k(\vec{q}_j)} \vec{m}_n,
\end{equation}
where $\mathcal{N}_k$ holds the indices of the $k$ nearest submap points $\vec{m}_n$ within a gating distance. Each match is uncertain through both the sonar beam and the submap, which we combine in
\begin{equation}
C_j = R\,B_j\,R^T + N_j + s_0^2\,I .
\end{equation}
Here, the beam covariance is rotated into the submap with the rotation of the current pose, $N_j$ is the spread of the matched submap points, and the floor $s_0$ keeps the weights finite. The squared Mahalanobis distance of every detection to its match is then
\begin{equation}
d_j^2 = (\vec{q}_j - \vec{\mu}_j)^T\, C_j^{-1}\, (\vec{q}_j - \vec{\mu}_j),
\end{equation}
and the correction minimizes these distances together with a motion prior:
\begin{equation}
\vec{\xi}^{*} = \arg\min_{\vec{\xi}} \; \sum_j w_j\,\rho\big(d_j^2\big) + \vec{\xi}^T \Sigma_p^{-1}\, \vec{\xi}.
\label{eq:ao}
\end{equation}
The weight $w_j$ is the square root of the strength of the detection relative to the strongest one, $\rho$ is a Cauchy loss that suppresses wrong associations, and $\Sigma_p$ is the covariance of the prior, discussed below. We solve (\ref{eq:ao}) by Gauss-Newton with re-association. This is a point-to-distribution registration, close to Generalized-ICP \cite{segalGeneralizedICP2009} and the normal distributions transform \cite{biberNormalDistributionsTransform2003}, but with the uncertainty of the sonar beam instead of a local surface model.

The prior predicts the rotation from an attitude filter and a translation along the body $x$-axis equal to the wheel distance since the previous capture. Its covariance $\Sigma_p$ (Table~\ref{tab:ao}) decides which degrees of freedom the sonar may change. The rotation follows the gyroscope: with a few hundred detections within a few meters, a small rotation is hard to tell apart from a small translation, and loosening the rotation prior from \SI{0.3}{\degree} to \SI{1}{\degree} per step increased the rotation error over \SI{20}{\meter} of travel from \SI{3.1}{\degree} to \SI{17}{\degree}, against \SI{2.4}{\degree} for the gyroscope alone (Section~\ref{sec:res_ao}). The horizontal translation follows the sonar, through a loose prior. The vertical translation follows the wheels and the IMU, as the height of a wheeled robot changes with its pitch and the distance driven. The sonar can estimate it too, but elevation is its weakest dimension, and its vertical estimate added a random walk to the height that the loop closures could not remove; the tight vertical prior halved the height error of the final trajectory (Section~\ref{sec:res_slam}).

The attitude filter is a complementary filter on $SO(3)$ \cite{mahonyNonlinearComplementaryFilters2008}. It integrates the gyroscope, with its bias estimated during a standstill at the start of the recording, and pulls the estimated gravity direction toward the measured one with a time constant of \SI{10}{\second}. While driving, the accelerometer also senses the motion of the robot, which we remove first:
\begin{equation}
\vec{a}_{\text{grav}} = \vec{a}_{\text{meas}} -
\begin{bmatrix} \dot{v} \\ v\,\omega_z \\ -v\,\omega_y \end{bmatrix},
\end{equation}
with $v$ the wheel speed and $\omega$ the body angular rate.

\begin{table}[t]
\centering
\caption{Parameters of the 3D Acoustic Odometry (Prior per Step; Forward Prior Grows by \SI{5}{\percent} of the Wheel Distance).}
\label{tab:ao}
\setlength{\tabcolsep}{4pt}
\begin{tabular}{lr}
\toprule
Submap & 20 captures \\
Beam (range / az. / el.) & \SI{3}{\centi\meter} / \SI{3}{\degree} / \SI{5}{\degree} \\
Neighbors / gate & 6 / \SI{0.6}{\meter} \\
Cov. floor / Cauchy scale & \SI{5}{\centi\meter} / 2.5 \\
\midrule
Prior, rotation & \SI{0.05}{\degree} \\
Prior, fwd. / lat. / vert. & \SI{1}{\centi\meter} / \SI{5}{\centi\meter} / \SI{5}{\milli\meter} \\
\bottomrule
\end{tabular}
\end{table}

\subsection{6-DoF Pose Graph}
\label{sec:posegraph}
After every \SI{1}{\meter} of travel, a keyframe $i$ with pose $X_i \in SE(3)$ is added to a factor graph that is optimized incrementally with iSAM2 \cite{kaessISAM2IncrementalSmoothing2012, dellaertFactorGraphsGTSAM2012}. Consecutive keyframes are connected by the acoustic odometry between them, with standard deviations that grow with the distance $\Delta s$ between the keyframes (\SI{2}{\centi\meter} plus \SI{5}{\percent} of $\Delta s$ in translation, \SI{0.15}{\degree} plus \SI{0.25}{\degree} per meter in rotation). Every keyframe also receives an attitude factor: with $\vec{a}_i$ the direction of gravity measured by the attitude filter in the body frame, the factor penalizes the angle between $R_i\cdot\vec{a}_i$ and the vertical $\vec{e}_z$ of the world (two degrees of freedom, standard deviation \SI{1.5}{\degree}). The attitude factors make roll and pitch absolute, so that the map stays level over long drives, whereas yaw and position are only constrained by the odometry and the loop closures.

\subsection{3D Place Recognition}
\label{sec:place}
A single sonar capture is a poor description of a place in a forest. We therefore describe a keyframe by its submap: the detections of the last \SI{4}{\meter} of travel, transformed with the acoustic odometry into the keyframe's level frame (its position and heading, without roll and pitch, which the IMU provides), and merged into \SI{0.1}{\meter} voxel means. Inspired by Scan Context \cite{kimScanContextEgocentric2018}, we summarize the submap in a polar descriptor $D[l,n]$ with rings $l$ of \SI{0.5}{\meter} from 1 to \SI{8}{\meter} and $60$ sectors $n$ of \SI{6}{\degree}, containing all submap points from just below the ground up to \SI{2.6}{\meter}. Every cell holds the summed $\sqrt{w_j}$ of the detections in it, compressed with an exponent $\gamma = 0.5$, and the descriptor is normalized to unit norm. The similarity of a query $q$ and a candidate $a$ is the best circular correlation over the sector shift $\Delta n$:
\begin{equation}
S(q,a) = \max_{\delta} \sum_{l,n} D^q_{l,n}\, D^a_{l,\,n+\delta},
\end{equation}
where $\delta$ is a shift by whole sectors and the sector index is taken modulo the number of sectors. The best shift in this map gives the relative heading of the two keyframes to within \SI{6}{\degree}. One could argue that a 3D sonar deserves a 3D descriptor, and we also tested four separate height bands (ground, 0.25--0.75, 0.75--1.5 and 1.5--\SI{2.6}{\meter}). In our forest, however, the height bands retrieved the correct place less often (section \ref{sec:res_place}), most likely because the height of a detection on a trunk depends on where the beam happens to hit it, so that the bands split the evidence of one object over several channels. Nevertheless, the 3D information is used where it matters most: in the leveled submap, and in the verification by registration.

In a strongly self-similar forest, no single descriptor will single out the right place from a database of hundreds of keyframes, and our results confirm this (section \ref{sec:res_place}). As in the 2D RTISSLAM on real data, we therefore search only among keyframes that lie within $\SI{5}{\meter}$ of the current position estimate and were visited at least \SI{20}{\meter} of travel earlier. The five best candidates by $S(q,a)$ are verified by a 4-DoF registration (position and yaw, as roll and pitch are known from the IMU) of the query submap onto the candidate submap, from two initial guesses: the descriptor's heading, and the relative pose of the current estimate. The registration uses the same point-to-distribution cost as the acoustic odometry, with a coarse-to-fine association distance (1.5 down to \SI{0.3}{\meter}). Ground points get a lower weight, because they fix the height, but do not contribute much to constraining the horizontal position. A candidate is accepted when at least \SI{85}{\percent} of the query's above-ground points have a candidate point within \SI{0.3}{\meter} after registration, and their RMS distance is at most \SI{0.12}{\meter}. The best accepted candidate becomes a loop closure factor between the two keyframes: the registered level-frame transform, combined with the roll and pitch of both keyframes, with standard deviations of \SI{0.10}{\meter} horizontally, \SI{0.05}{\meter} vertically, \SI{1}{\degree} in yaw and \SI{3}{\degree} in roll and pitch, and a Cauchy kernel that limits the influence of a wrong closure \cite{agarwalRobustMapOptimization2013}.

\subsection{3D Occupancy Map}
\label{sec:map}
The map is a log-odds occupancy grid with \SI{0.1}{\meter} voxels \cite{elfesUsingOccupancyGrids1989, hornungOctoMapEfficientProbabilistic2013}, built on the GPU from the detections of all captures. Each capture is placed relative to the keyframe before it with the acoustic odometry, so the map follows the optimized keyframe poses.

Given its beamwidth, a sonar detection is not a point but a patch of the beam at the measured range. Each detection therefore spreads its occupied evidence over 24 samples of the same beam-shaped Gaussian as in the odometry: thin along the range ray and wide across it. The evidence scales with the strength of the detection, from 0.4 to 1 of the full increment of 4 over a \SI{25}{\decibel} range. A single detection thus smears over the beam, but views from many positions concentrate the evidence on the real surfaces.

Free space is carved along each direction from \SI{0.5}{\meter} up to \SI{0.15}{\meter} before the first echo in that direction, with a decrement of 0.12 per voxel and capture. Directions without an echo are carved only up to \SI{1}{\meter}. A specular surface at grazing incidence returns no echo, so an empty direction does not prove that the space in it is empty. The log-odds are clipped to the interval from $-2$ to $5$, and voxels with a log-odds of at least 2 are occupied.
\section{Experimental Setup}
\label{sec:setup}
In this section, we describe the robot and the recording, the reference trajectory and map against which we evaluate RTISSLAM-3D, and the metrics we use.

\subsection{Robot and Recording}
The data were recorded with a Clearpath Husky A200 that carried an eRTIS 3D imaging sonar \cite{kerstensERTISFullyEmbedded2019, Laurijssen2026eRTIS} at the front, about \SI{0.32}{\meter} above the ground, together with an Ouster OS1 64 beam 3D lidar with a built-in 6-axis IMU (\SI{100}{\hertz}), the wheel encoders (\SI{11.5}{\hertz}), and a GoPro HERO13 camera (which we only use for illustration). The eRTIS has $N = 32$ MEMS microphones in a planar array with an aperture of \SI{66.5}{\milli\meter} by \SI{37.6}{\milli\meter}, and it emitted a \SI{2.5}{\milli\second} frequency-modulated call between \SI{25}{\kilo\hertz} and \SI{70}{\kilo\hertz} at \SI{4.34}{\hertz}, recording \SI{10}{\meter} of echoes at \SI{450}{\kilo\hertz} per pulse. The robot drove about \SI{560}{\meter} in \SI{23.4}{\minute} through a forest in Trondheim, Norway on 13 October 2026 under non-adverse weather conditions, on and next to forest paths, with several loops that were driven repeatedly in both directions. The recording holds 6090 sonar captures and 13941 lidar scans. All sensors were stored in a single HDF5 file with a common robot clock. 

\subsection{Reference}
As reference, we use a 3D lidar SLAM of the same drive, computed with KISS-SLAM \cite{guadagninoKISSSLAMSimpleRobust2025}, which combines KISS-ICP odometry \cite{vizzoKISSICPDefensePointtoPoint2023} with loop closures between local maps and a pose-graph optimization. We checked the reference in three ways: its horizontal trajectory agrees with a 2D lidar SLAM (slam\_toolbox \cite{macenskiSLAMToolboxSLAM2021}) to within \SI{0.35}{\meter} RMS, its height at 844 revisits of the same place is consistent to within \SI{3.4}{\centi\meter} RMS, and its roll and pitch agree with our IMU attitude filter to within \SI{1.2}{\degree} and \SI{1.0}{\degree} RMS respectively. A lidar SLAM system, however, has no knowledge about the direciton of the gravity vector, as its world frame is that of its first scan. When we aligned it with the attitude of the IMU at the start of the drive only, a residual tilt of \SI{1.5}{\degree} remained, which showed up as a height error that grew linearly with the horizontal position. This is not a negligible effect: over a site of \SI{80}{\meter}, it amounts to \SI{2}{\meter} of height gain that is only due to a miscalibration of the world reference frames. We therefore align the reference with gravity using the whole drive: we rotate the motion-corrected accelerometer into the reference's world frame with the reference's own attitude, average it over the whole drive (the driving accelerations average out), and rotate the reference so that this average points straight up. 

\subsection{Metrics}
We compare every keyframe pose with the reference pose at the same time. Because gravity is known to both, we align the trajectories with a 4-DoF fit (translation and yaw), and report the horizontal ATE (Absolute Trajectory Error, RMS), the RMS height error and the maximum horizontal error. We also report the RMS roll and pitch errors, and the tilt that a full 6-DoF fit would add, which measures how level the estimated map is. For the odometry, we report the RPE (Relative Pose Error) over segments of 5 and \SI{20}{\meter} of travel. Every accepted loop closure is checked against the relative pose of the two keyframes in the reference. The 3D map is compared with the lidar map (\SI{0.2}{\meter} voxels) after moving it into the lidar frame with a 6-DoF fit of the trajectories: the precision is the fraction of occupied sonar voxels within \SI{0.3}{\meter} of a lidar voxel, and the recall the fraction of lidar voxels within the volume observed by the sonar (within \SI{6}{\meter} in front of a sensor pose) that have a sonar voxel within \SI{0.3}{\meter}. Both are given per height band above the local ground: the ground (below \SI{0.25}{\meter}), trunks and vegetation (0.25--\SI{2}{\meter}), and the canopy (above \SI{2}{\meter}). The complete pipeline (imaging, detection, odometry, SLAM, mapping and the videos) runs with a single command on a workstation with an NVIDIA RTX PRO 5000 GPU, in Python with PyTorch 2.14, GTSAM 4.2 and SciPy.

\section{Results}
\label{sec:results}
In this section, we will first look at the 3D detections of single captures. Next, we evaluate the 3D acoustic odometry and the 3D place recognition on their own, after which we present the trajectory of the complete system together with an ablation of its components and a sensitivity analysis of its thresholds. We conclude with the 3D map and the computation times.

\subsection{3D Imaging and Detections}
\label{sec:res_imaging}
\begin{figure*}[t]
\centering
\includegraphics[width=\linewidth]{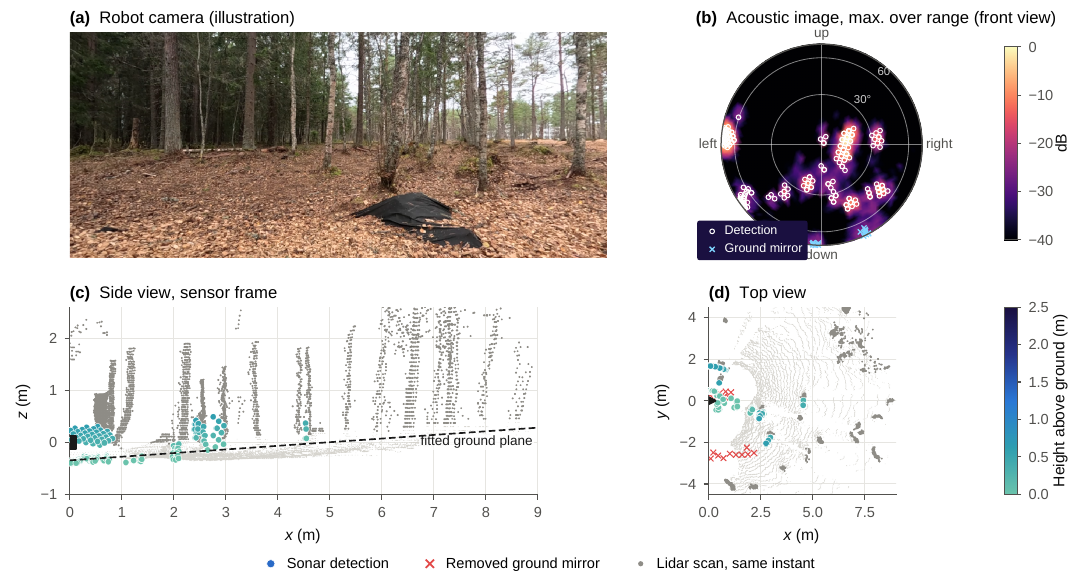}
\caption{3D Detections of a Single eRTIS Capture in the Forest. Panel a) shows the view of the robot's camera (for illustration only). Panel b) shows the acoustic image of the capture (DMAS5 with coherence factor, maximum over range, \SI{40}{\decibel} dynamic range) over the frontal hemisphere as seen from behind the sensor, with circles at \SI{30}{\degree} and \SI{60}{\degree} off the sensor axis, and the directions of the CFAR detections (white circles; cyan crosses: ground mirrors). Panels c) and d) show the 3D detections in the sensor frame, in a side view and a top view, colored by their height above the fitted ground plane, on top of the lidar scan of the same instant (dark gray: lidar returns above the ground, light gray: ground). Red crosses: detections removed as ground multipath mirrors, i.e., more than \SI{0.15}{\meter} below the fitted ground plane (dashed line in panel c)). The sonar sees the ground in front of the robot and the near sides of the trunks.}
\label{fig:imaging}
\end{figure*}

Figure \ref{fig:imaging} shows a typical capture, taken on a forest path with trunks at 2 to \SI{5}{\meter}, next to the lidar scan of the same moment. The sonar detections lie on the ground in front of the robot and on the near sides of the trunks, where the lidar also sees them. The removed detections lie below the fitted ground plane, where the mirror images of the ground multipath appear. Over the whole recording, the detector found a median of 160 detections per capture, of which \SI{18}{\percent} were removed as ground mirrors. A ground plane was found in \SI{92}{\percent} of the captures, and the median height of the sensor above it was estimated to be \SI{0.316}{\meter} (\SI{10}{\percent} to \SI{90}{\percent} of the captures between \SI{0.27}{\meter} and \SI{0.37}{\meter}). This agrees with the mounting height: the extrinsic calibration of the platform places the sonar \SI{0.158}{\meter} above the wheel axles, which, with the wheel radius of \SI{0.165}{\meter} of the Husky, puts it \SI{0.32}{\meter} above flat ground. It should be noted that, as seen from the sonar, the lidar points of this recording only cover elevations between \SI{-8}{\degree} and \SI{22}{\degree}, so that the lidar does not see the ground close to the robot, whereas the sonar sensor does.

\subsection{3D Acoustic Odometry}
\label{sec:res_ao}
\begin{figure*}[t]
\centering
\includegraphics[width=\linewidth]{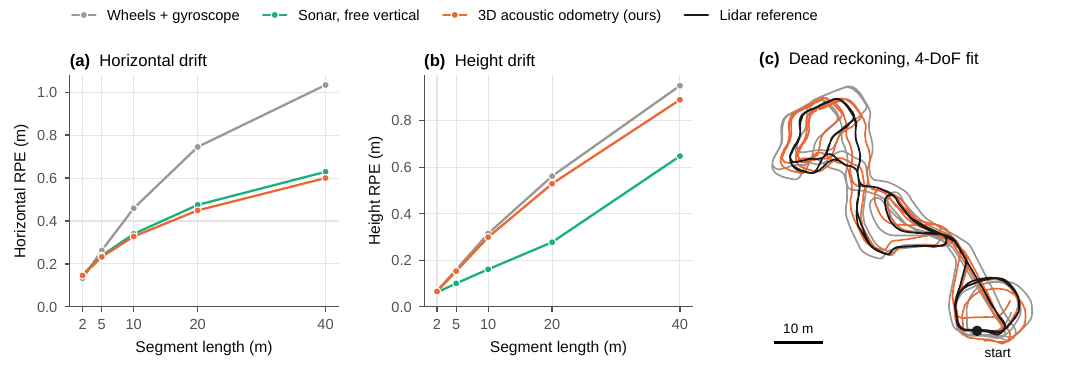}
\caption{3D Acoustic Odometry Against the Lidar Reference. Panel a) shows the horizontal RPE (Relative Pose Error) as a function of the segment length, for the wheels and gyroscope alone (gray), the acoustic odometry in which the sonar estimates the full translation (green), and the final 3D acoustic odometry, in which the sonar estimates the horizontal translation only (orange). Panel b) shows the height RPE. Panel c) shows the dead-reckoning trajectories of the wheels and gyroscope and of the 3D acoustic odometry after a 4-DoF fit on the lidar reference (black). The sonar reduces the horizontal drift of the wheels by about \SI{40}{\percent}, equally in both variants; estimating the vertical motion with the sonar reduces the height drift of the odometry alone, but not the error of the final SLAM trajectory (see text).}
\label{fig:odometry}
\end{figure*}

Figure~\ref{fig:odometry} compares three odometry variants against the reference: the wheels and the gyroscope alone (the prior of the acoustic odometry), the acoustic odometry in which the sonar estimates the full translation, and the final 3D acoustic odometry, in which the sonar estimates only the horizontal translation and the height follows the wheels and the IMU. Horizontally, the sonar reduces the drift of the wheels by about \SI{40}{\percent}, and both sonar variants perform alike: over \SI{20}{\meter} of travel, the horizontal RPE drops from \SI{0.74}{\meter} to \SI{0.47}{\meter} with the full translation and \SI{0.45}{\meter} for the final variant. The variants differ only in height. On its own, the sonar's vertical estimate drifts less than the wheels and the IMU (a height RPE of \SI{0.28}{\meter} against \SI{0.53}{\meter} over \SI{20}{\meter}), but inside the SLAM it gives the worse result: fewer verified loop closures (217 against 233), and a larger height error (\SI{0.23}{\meter} against \SI{0.12}{\meter}) and horizontal error (\SI{0.32}{\meter} against \SI{0.28}{\meter}) of the final trajectory (Table~\ref{tab:slam}). We attribute this to the nature of the errors: the height drift of the wheels and the IMU is slow, as it stems from small pitch errors, and is removed by the loop closures, whereas the vertical estimate of the sonar rests on its weakest dimension, elevation, and disturbs the submaps on which the loop closures are verified. We therefore let the sonar estimate the horizontal translation only, as rotation is best left to the gyroscope as well (Section~\ref{sec:ao}).

\subsection{3D Place Recognition}
\label{sec:res_place}
To evaluate the place recognition on its own, we built a keyframe every \SI{1}{\meter} along the acoustic odometry (533 keyframes), and compared every keyframe with all keyframes at least \SI{20}{\meter} of travel earlier, without any position gate. Here, 367 keyframes had a true revisit (a keyframe within \SI{2}{\meter}) in that set. The descriptor alone ranked a true revisit first for \SI{68}{\percent} of them, and among the top five for \SI{79}{\percent}. The descriptor with four height bands did worse, with \SI{56}{\percent} and \SI{71}{\percent}. As expected for a forest, the descriptor is thus useful to rank candidates, but by itself it is far from reliable enough to close loops.

\begin{figure*}[t]
\centering
\includegraphics[width=\linewidth]{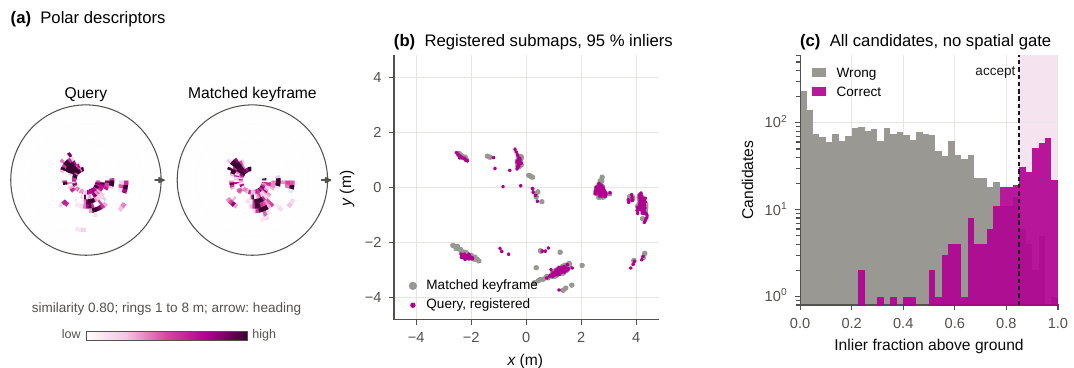}
\caption{3D Place Recognition. Panel a) shows the polar descriptors (rings from 1 to \SI{8}{\meter}, 60 sectors, the arrow marks the heading) of a query keyframe and of the keyframe it was matched with, for a loop closure with the median error. Panel b) shows the above-ground points of the two keyframe submaps after the 4-DoF registration (gray: matched keyframe, magenta: query). Panel c) shows the distribution (logarithmic scale) of the inlier fraction of the above-ground points after registration, for all top-five candidates of all keyframes without a position gate, split into candidates that were registered to within \SI{0.5}{\meter} and \SI{5}{\degree} of the reference (magenta) and the others (gray); the dashed line is the acceptance threshold.}
\label{fig:place}
\end{figure*}

The verification by registration, however, separates the correct candidates from the wrong ones very well (figure \ref{fig:place}c). Of the top-five candidates of all keyframes, again without a position gate, 264 passed our acceptance test (\SI{85}{\percent} of the above-ground points within \SI{0.3}{\meter}, RMS distance at most \SI{0.12}{\meter}), and 256 of those (\SI{97}{\percent}) were registered to within \SI{0.5}{\meter} and \SI{5}{\degree} of the reference. The remaining eight were all 4 to \SI{38}{\meter} away from the query, and are therefore excluded by the position gate in the SLAM system. Figure \ref{fig:place} also shows a typical example: the descriptors of a query and its accepted candidate, rotated by the shift that the descriptor estimated, and the two submaps after registration.

\subsection{6-DoF Trajectory}
\label{sec:res_slam}
\begin{figure*}[t]
\centering
\includegraphics[width=\linewidth]{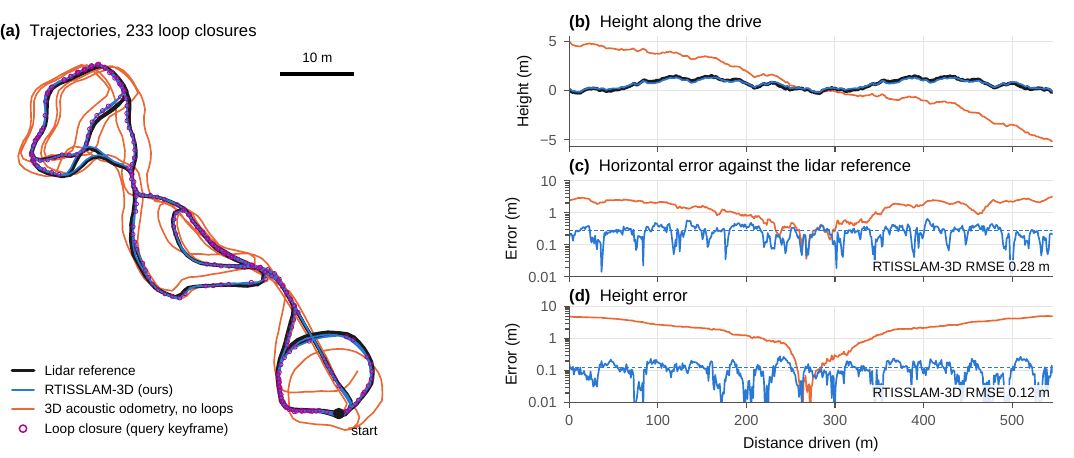}
\caption{RTISSLAM-3D Against the 3D Lidar Reference. Panel a) shows the trajectories after a 4-DoF fit: the reference (black), the 3D acoustic odometry without loop closures (orange) and RTISSLAM-3D (blue); magenta circles mark the query keyframes of its 233 loop closures. Panel b) shows the height along the drive. Panels c) and d) show the horizontal and the height error of both trajectories on a logarithmic scale; the dashed lines are the RMSE of RTISSLAM-3D.}
\label{fig:trajectory}
\end{figure*}

Figure \ref{fig:trajectory} shows the trajectory of RTISSLAM-3D next to the 3D acoustic odometry and the reference, and table \ref{tab:slam} lists the errors. With 233 loop closures, the horizontal ATE drops from \SI{1.74}{\meter} for the acoustic odometry to \SI{0.28}{\meter}, with a maximum error of \SI{0.64}{\meter}, and the height error drops from \SI{2.82}{\meter} to \SI{0.12}{\meter}. The roll and pitch errors (\SI{1.2}{\degree} and \SI{1.0}{\degree}) are those of the IMU attitude filter, which the loop closures neither improve nor degrade, and the estimated map is level to within \SI{0.34}{\degree}. None of the 233 loop closures is off by more than \SI{0.5}{\meter} against the reference: their median error is \SI{7.5}{\centi\meter} horizontally, \SI{2.6}{\centi\meter} in height and \SI{0.57}{\degree} in yaw. For comparison, the 2D RTISSLAM on the same recording, lifted to 3D with the same IMU attitude and with its heights made consistent at its loop closures, reaches a horizontal error of \SI{0.68}{\meter} and a height error of \SI{0.94}{\meter} against the same reference: its map is tilted by \SI{4.4}{\degree}, because a planar tilt of the whole map cannot be observed from loop closures alone. In RTISSLAM-3D, the attitude factors keep the map level.

\subsection{Ablation and Sensitivity}
\label{sec:res_ablation}
\begin{table}[t]
\centering
\caption{Ablation of RTISSLAM-3D Against the Lidar Reference: Horizontal ATE, Height RMSE and Maximum Horizontal Error After a 4-DoF Fit (m), and Loop Closures With an Error Above \SI{0.5}{\meter} (Bad).}
\label{tab:slam}
\setlength{\tabcolsep}{4pt}
\begin{tabular}{lrrrrr}
\toprule
Variant & Loops & $xy$ & $z$ & Max & Bad \\
\midrule
\textbf{RTISSLAM-3D (ours)} & \textbf{233} & \textbf{0.28} & \textbf{0.12} & \textbf{0.64} & \textbf{0} \\
No loop closures & 0 & 1.74 & 2.82 & 3.23 & -- \\
Wheel odometry & 237 & 0.81 & 0.09 & 1.46 & 0 \\
Wheel odometry, no loops & 0 & 2.40 & 3.00 & 3.87 & -- \\
Sonar vertical & 217 & 0.32 & 0.23 & 0.69 & 0 \\
Four height bands & 236 & 0.28 & 0.12 & 0.66 & 0 \\
\bottomrule
\end{tabular}
\end{table}

Table \ref{tab:slam} shows what every component contributes. Without loop closures, the acoustic odometry ends up with \SI{1.74}{\meter} of horizontal error; with loop closures but with the wheels and the gyroscope as odometry, the horizontal error is \SI{0.81}{\meter}, so that the sonar odometry and the loop closures are both needed, and they complement each other. Interestingly, the wheels and the gyroscope with loop closures give the lowest height error of all variants (\SI{0.09}{\meter}), which again shows that the height of a wheeled robot is best left to its pitch and its wheels. Letting the sonar also estimate the vertical motion doubles the height error, as discussed in section \ref{sec:res_ao}, and the descriptor with height bands gives the same trajectory as the one without, because the registration, not the descriptor, decides which loop closures are accepted.

The result is insensitive to the thresholds of the loop closure test (Appendix~\ref{app:thresholds}, Table~\ref{tab:thresholds}): across inlier thresholds from 0.75 to 0.95 and position gates from 3 to \SI{12}{\meter}, the number of loop closures varies from 95 to 290, but the horizontal ATE stays between 0.27 and \SI{0.29}{\meter}, and at most one loop closure had an error above \SI{0.5}{\meter}, which the robust kernel absorbed. We are therefore confident that the result is not a consequence of a fortunate choice of thresholds, although only a second recording can show this conclusively.

\subsection{3D Map}
\label{sec:res_map}
\begin{figure*}[t]
\centering
\includegraphics[width=\linewidth]{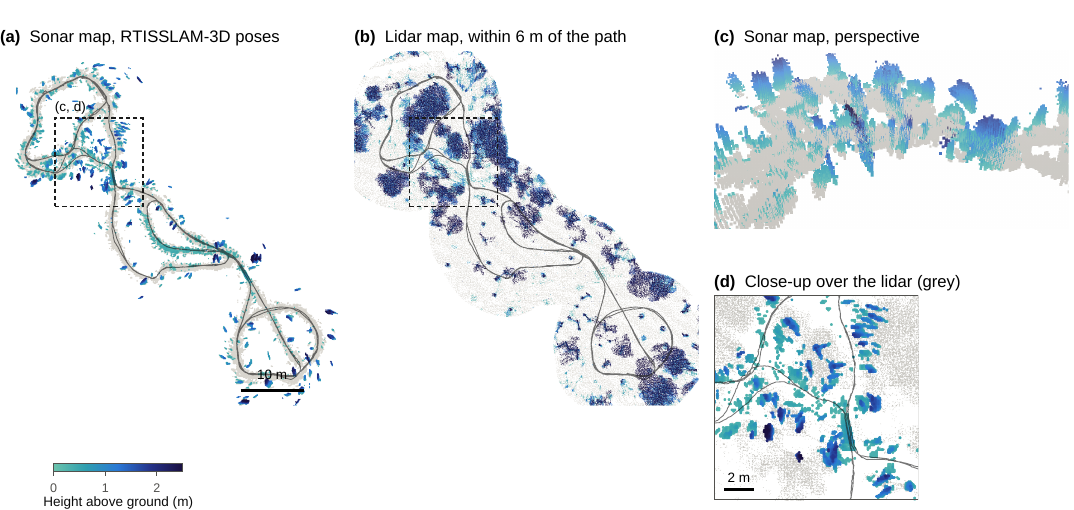}
\caption{3D Map of the Forest. Panel a) shows the 3D occupancy map of RTISSLAM-3D (\SI{0.1}{\meter} voxels) from above and panel b) the lidar map of the reference, limited to the points within \SI{6}{\meter} in front of a sensor pose (the volume in which the recall is scored, Section~\ref{sec:setup}) and up to \SI{3}{\meter} above the ground, at the same scale; the gray line is the path of the robot. Colors encode the height above the local ground; the ground itself is light gray. Panel c) shows the sonar map of the most revisited part of the site (dashed square) in perspective, and panel d) the same part from above, the sonar map above the ground drawn over the lidar map above the ground (gray). The trunks and the vegetation along the paths are resolved in the sonar map; the canopy is outside the sonar's view.}
\label{fig:map}
\end{figure*}

Figure \ref{fig:map} shows the 3D occupancy map next to the lidar map, and table \ref{tab:map} lists the precision and recall per height band. Of the occupied sonar voxels, \SI{94.6}{\percent} lie within \SI{0.3}{\meter} of a lidar voxel, and \SI{92.7}{\percent} of those on trunks and vegetation between 0.25 and \SI{2}{\meter}. The same map built on the poses of the lidar reference, which is the upper bound for any sonar SLAM with these detections, reaches \SI{96.0}{\percent} and \SI{94.1}{\percent}: the pose errors of RTISSLAM-3D cost only about one percentage point of precision. The recall is low by nature: the sonar sees the ground in front of the robot and the near sides of the trunks within a few meters, whereas the lidar sees the trunks from all sides and up to the canopy, which the sonar does not see at all.

\begin{table}[t]
\centering
\caption{Precision and Recall of the 3D Sonar Map Against the Lidar Map, per Height Band (Tolerance \SI{0.3}{\meter}; Recall Within \SI{6}{\meter} in Front of the Sensor Poses).}
\label{tab:map}
\begin{tabular}{lrrr}
\toprule
 & \multicolumn{2}{c}{Precision $\uparrow$} & Recall $\uparrow$ \\
\cmidrule(lr){2-3}
Height Band & Lidar Poses & \textbf{Ours} & \textbf{Ours} \\
\midrule
Ground ($<$\,\SI{0.25}{\meter}) & 0.970 & \textbf{0.959} & 0.363 \\
Trunks, Vegetation (0.25--\SI{2}{\meter}) & 0.941 & \textbf{0.927} & 0.243 \\
Canopy ($>$\,\SI{2}{\meter}) & 0.977 & \textbf{0.938} & 0.010 \\
\midrule
All & 0.960 & \textbf{0.946} & 0.141 \\
\bottomrule
\end{tabular}
\end{table}

\subsection{Computation Time}
\label{sec:res_runtime}
On our workstation, the 3D imaging of all 6090 captures took \SI{471}{\second} on the GPU (\SI{77}{\milli\second} per capture, against a pulse interval of \SI{230}{\milli\second}), the detection \SI{45}{\second}, the 3D acoustic odometry of 6043 captures \SI{238}{\second} and the SLAM back-end with the place recognition \SI{236}{\second} (both on a single CPU thread, in Python), and the 3D map \SI{29}{\second}. For a drive of \SI{23.2}{\minute}, the whole chain thus runs about 1.4 times faster than real time.

\begin{figure*}[t]
\centering
\includegraphics[width=\linewidth]{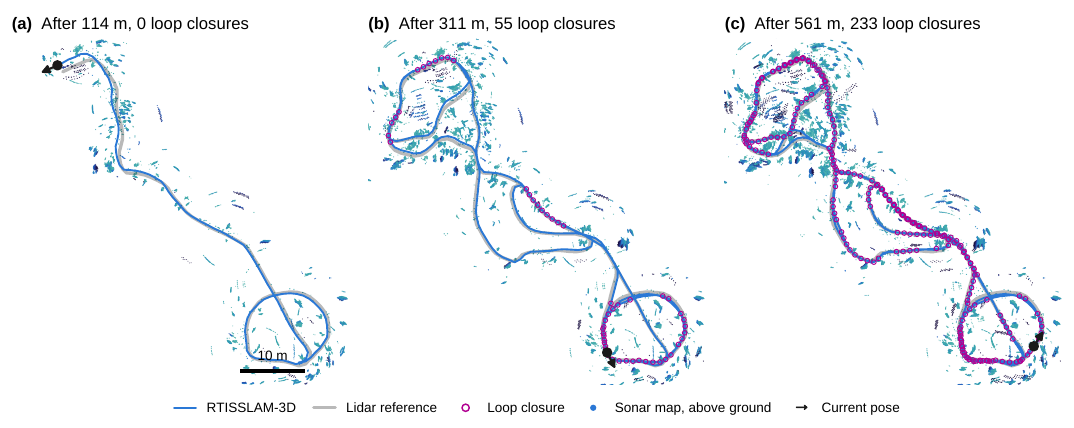}
\caption{The Map and Trajectory as RTISSLAM-3D Builds Them (Three Moments of the Supplementary Video, Top View, Same Scale). The sonar map above the ground is carried by the keyframe poses of the estimate at that moment, the trajectory of RTISSLAM-3D is drawn in blue and the lidar reference in gray; magenta circles mark the loop closures found so far and the arrow the current pose. Panel a) shows the first lap, before any loop closure, panel b) the middle of the drive, and panel c) the final map.}
\label{fig:build}
\end{figure*}

The supplementary video (\url{https://youtu.be/GHdZd9Brqog}) shows the SLAM system at work: the keyframe poses appear one by one, the map grows with them, and the trajectory snaps into place whenever loop closures are added (figure \ref{fig:build}).

\section{Discussion and Limitations}
\label{sec:discussion}
The results of the previous section show that an in-air 3D imaging sonar, an IMU and the wheels are enough to estimate the 6-DoF trajectory of a robot in a forest with an error of a few decimeters, and to build a usable 3D occupancy map in the process. In what follows, we discuss what made this possible, and where the limits of the current system lie.

While showing promising results, the RTISSLAM-3D has some limitations. For one, we evaluated RTISSLAM-3D on a single recording, which was also used to choose its parameters. The sensitivity analysis in section \ref{sec:res_ablation} shows that the trajectory hardly depends on the thresholds of the loop closure test, but more extensive tests in a wider set of environments is needed in future work.

Another shortcoming is that the reference is itself the output of a SLAM system (based on Lidar). We have shown that its world frame was tilted by \SI{1.5}{\degree} with respect to gravity when aligned in the usual way, which would have appeared as a height error of RTISSLAM-3D of more than half a meter, and we corrected it with the accelerometer data. The remaining errors of the reference are of the order of a few centimeters (its height at revisits is consistent to within \SI{3.4}{\centi\meter}), well below the errors we report. While it is true that comparing the outputs of SLAM systems has risks, it is the only practical approach to perform large-scale real-world evaluations of SLAM systems, without resorting to simulation environments.

Furthermore, the loop closures are only searched within \SI{5}{\meter} of the current estimate. This works as long as the odometry drifts less than that between revisits, which was the case here (the acoustic odometry drifted at most \SI{3.2}{\meter} over the whole drive). For larger loops, the gate can be widened (the result was unchanged up to \SI{12}{\meter}), and beyond that, the high precision of the ungated verification suggests a global search with a consistency check over several loop closures.

The attitude is taken from a complementary filter, and only its gravity direction enters the pose graph as an absolute factor. A tighter integration of the IMU through preintegrated factors \cite{forsterOnManifoldPreintegration2017} would be the natural next step, and would also allow the sonar to contribute to the rotation where it is well constrained. Likewise, the tight vertical prior assumes a wheeled robot that moves along its body $x$-axis. On a legged or flying robot, the vertical motion would have to come from the sonar, or from another sensor.

The map has a low recall by nature, as the sonar only sees the near sides of objects within a few meters, and does not see the canopy at all. This is not a limitation for navigation close to the ground, which is what the sonar is for, but it means that the sonar map is a different kind of map than a lidar map, and should be judged as such.

A final shortcoming is that the current implementation processes a recording offline, in Python. The imaging runs at \SI{77}{\milli\second} per capture on the GPU, well within the pulse interval of \SI{230}{\milli\second}, and the odometry and the SLAM back-end together run about three times faster than real time on a single CPU thread in Python (section \ref{sec:res_runtime}), so that an online implementation, as for the 2D RTISSLAM in ROS 2, is a matter of engineering rather than of research.

\section{Conclusion}
\label{sec:conclusion}
In this paper, we presented RTISSLAM-3D, a novel full 6-DoF SLAM system for an in-air 3D imaging sonar on a wheeled robot, which uses the sonar's 3D images for odometry, for place recognition and for the metric verification of loop closures, complemented only by an IMU and the wheel encoders. Every capture is imaged over the frontal hemisphere and reduced to a sparse 3D detection set by a CFAR detector on the sphere, with the ground multipath removed through a fitted ground plane. A 3D acoustic odometry registers these detections with their beam-shaped uncertainty, keyframes enter a 6-DoF pose graph with IMU attitude factors, and loop closures are proposed by a polar descriptor of the keyframe submaps and verified by 4-DoF submap registration. On a \SI{545}{\meter} drive through a forest, against a gravity-aligned 3D lidar reference, RTISSLAM-3D reduced the horizontal trajectory error from \SI{2.40}{\meter} for the wheels and the gyroscope and \SI{1.74}{\meter} for the 3D acoustic odometry to \SI{0.28}{\meter}, with a height error of \SI{0.12}{\meter}, and none of its 233 loop closures was off by more than \SI{0.5}{\meter}. Its 3D map reached a precision of \SI{94.6}{\percent} against the lidar map, against \SI{96.0}{\percent} for the same map on the lidar poses.

We believe that with this paper, in-air sonar SLAM takes the step from planar SLAM to true 3D SLAM, and from simulation to the field. In future research, we will evaluate RTISSLAM-3D on further environments with frozen parameters, integrate the IMU more tightly into the pose graph, and port the system to ROS 2 for online use. The complete pipeline, from the raw recording to the map and the videos, will be made available as open source.

\FloatBarrier
\appendices
\section{Sensitivity to the Loop Closure Test}
\label{app:thresholds}
Table~\ref{tab:thresholds} repeats the full RTISSLAM-3D run with one threshold of the loop closure test changed at a time: the acceptance threshold on the inlier fraction of the above-ground points after registration, the radius of the position gate around the predicted pose, and the upper limit on the registration RMSE (\SI{0.15}{\meter} instead of \SI{0.12}{\meter}).

\begin{table}[t]
\centering
\caption{Sensitivity of RTISSLAM-3D to the Loop Closure Test (All Else as the Default, in Bold; Errors as in Table~\ref{tab:slam}).}
\label{tab:thresholds}
\setlength{\tabcolsep}{4pt}
\begin{tabular}{lrrrrr}
\toprule
Test & Loops & $xy$ & $z$ & Max & Bad \\
\midrule
Inliers $\geq$ 0.75 & 290 & 0.27 & 0.13 & 0.62 & 1 \\
Inliers $\geq$ 0.80 & 259 & 0.27 & 0.12 & 0.64 & 0 \\
\textbf{Inliers $\geq$ 0.85} & \textbf{233} & \textbf{0.28} & \textbf{0.12} & \textbf{0.64} & \textbf{0} \\
Inliers $\geq$ 0.90 & 182 & 0.28 & 0.12 & 0.65 & 0 \\
Inliers $\geq$ 0.95 & 95 & 0.28 & 0.11 & 0.67 & 0 \\
Gate \SI{3}{\meter} & 248 & 0.28 & 0.12 & 0.65 & 0 \\
Gate \SI{8}{\meter} & 222 & 0.28 & 0.12 & 0.65 & 0 \\
Gate \SI{12}{\meter} & 214 & 0.28 & 0.12 & 0.63 & 0 \\
RMSE $\leq$ \SI{0.15}{\meter} & 237 & 0.29 & 0.09 & 0.68 & 1 \\
\bottomrule
\end{tabular}
\end{table}

\section*{Acknowledgment}
We disclose that the AI assistant Claude (Anthropic) was used in this work: to implement the imaging, odometry, place recognition, SLAM and mapping software, to run the experiments and the evaluation against the lidar reference, to generate the figures, the tables and the supplementary video, and to draft parts of the text. The authors directed the work, checked all results, reviewed and edited the text, and take full responsibility for the content of this article. The AI assistant is not an author of this work.

\FloatBarrier
\bibliographystyle{IEEEtran}
\bibliography{references,in_air_sonar_2024_2026,rtisslam3d}

\end{document}